\pdfoutput=1

\PassOptionsToPackage{x11names}{xcolor}
\documentclass[11pt]{article}

\usepackage[]{latex/acl}

\usepackage{times}
\usepackage{latexsym}
\usepackage{natbib}
\usepackage{kotex}
\usepackage{algorithm}
\usepackage{algorithmicx}
\usepackage[noend]{algpseudocode}
\usepackage{amssymb}
\usepackage{amsmath, amsfonts, amsthm}
\usepackage{dsfont}
\usepackage{mathtools}
\usepackage{amsthm}
\usepackage{booktabs} 
\usepackage{multirow}
\usepackage{subcaption}
\usepackage{adjustbox}
\usepackage{algpseudocode}
\usepackage[bottom, symbol]{footmisc}
\usepackage{hyperref}
\usepackage{empheq}
\usepackage{colortbl}

\newcommand{\argmax}{\mathop{\mathrm{argmax}}} 

\newcommand{\redc}{\textcolor{red}} 

\definecolor{light-gray}{gray}{0.9}

\newcommand{\xx}{\mathbf{x}}
\newcommand{\yy}{\mathbf{y}}

\newcommand{\ie}{\textit{i}.\textit{e}., }
\newcommand{\eg}{\textit{e}.\textit{g}., }

\usepackage[T1]{fontenc}

\usepackage[utf8]{inputenc}

\usepackage{microtype}

\usepackage{inconsolata}

\usepackage{soul}

\title{Model-based Preference Optimization\\in Abstractive Summarization without Human Feedback}

\author{
  Jaepill Choi\footnotemark[1] \;\; {\bf Kyubyung Chae}\footnotemark[1] \;\; {\bf Jiwoo Song} \;\; {\bf Yohan Jo} \;\; {\bf Taesup Kim}\footnotemark[2] \\
    Graduate School of Data Science, Seoul National University \\
  \texttt{\{jaepill9205, kyubyung.chae, sjiwoo, yohan.jo, taesup.kim\}@snu.ac.kr}
}

\begin{document}
\maketitle

\begin{abstract}

In abstractive summarization, the challenge of producing concise and accurate summaries arises from the vast amount of information contained in the source document. Consequently, although Large Language Models (LLMs) can generate fluent text, they often introduce inaccuracies by hallucinating content not found in the original source. While supervised fine-tuning methods that maximize likelihood contribute to this issue, they do not consistently enhance the faithfulness of the summaries. Preference-based optimization methods, such as Direct Preference Optimization (DPO), can further refine the model to align with human preferences. However, these methods still heavily depend on costly human feedback. In this work, we introduce a novel and straightforward approach called Model-based Preference Optimization (MPO) to fine-tune LLMs for improved summarization abilities without any human feedback. By leveraging the model’s inherent summarization capabilities, we create a preference dataset that is fully generated by the model using different decoding strategies. Our experiments on standard summarization datasets and various metrics demonstrate that our proposed MPO significantly enhances the quality of generated summaries without relying on human feedback. The code is publicly available at \url{https://github.com/cjaep/MPO}.

\end{abstract}

\footnotetext[1]{Equal contribution.}
\footnotetext[2]{Corresponding author.}

\section{Introduction}
\label{sec:intro}

Large Language Models (LLMs) have demonstrated remarkable capabilities in generating fluent and plausible text~\cite{gpt-j, llama1, mistral}.
However, despite these advancements, LLMs often produce summaries that, while plausible, contain incorrect or contradictory information—a phenomenon known as \textit{hallucination} \cite{maynez-etal-2020-faithfulness}. 
The fundamental reason for this issue is that LLMs are primarily trained to predict the most likely next token based on maximum likelihood, which is the most common objective for pre-training language models~\cite{king2022dont}. 
In principle, reinforcement learning based objectives can circumvent these failures by choosing an appropriate reward function~\cite{paulus2018a, tian2024finetuning}.
Recently, reinforcement learning from human feedback (RLHF) has focused on aligning language models with human preferences, thereby effectively enhancing the models’ summarization abilities~\cite{bohm-etal-2019-better, pasunuru-bansal-2018-multi, NEURIPS2020_1f89885d, paulus2018a, ramamurthy2023is}. 

\begin{figure}[t]
    \centering
    \includegraphics[width=1.0\columnwidth]{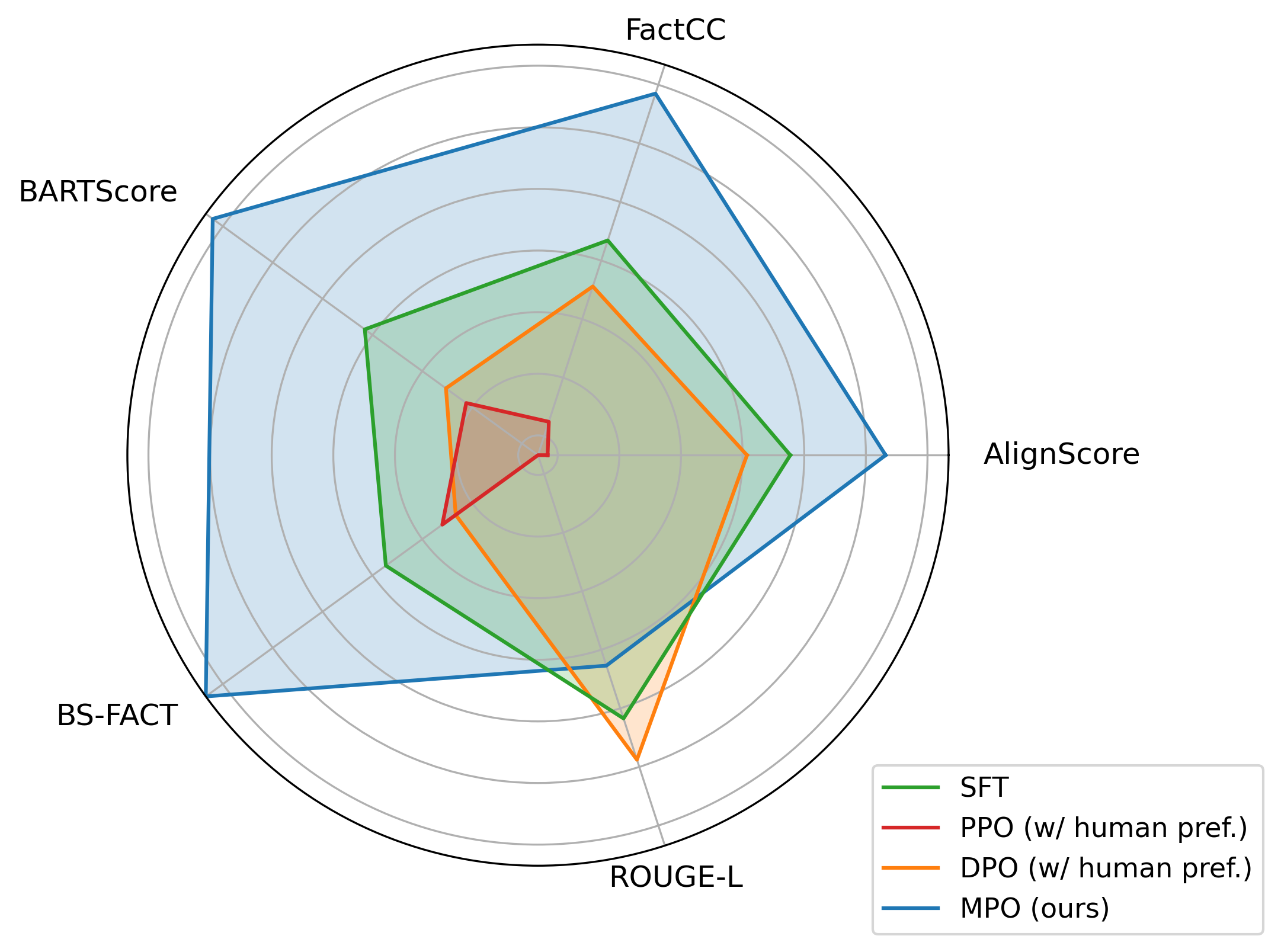}
    \caption{\textbf{Summarized results via automated metrics.} Our method MPO, which uses the model-generated summaries for preference optimization, proves to be more effective than PPO and DPO, both of which use human preference datasets for optimization. The results are from using the GPT-J on the TL;DR dataset.}
    \label{fig:radar_chart}
\end{figure}

While RLHF and other preference-based optimization methods~\cite{dpo} effectively fine-tune models to align with human preferences, human feedback is not always reliable.
For example, even though the quality of text summaries depends on various factors, \citet{hosking2024human} demonstrated that human preferences often overlook factuality and consistency, which are crucial in avoiding hallucination. 
This implies that a summary judged as favorable by humans is not necessarily free from hallucination. 
In other words, preference optimization with human feedback does not guarantee improved faithfulness.
Moreover, the use of human preference faces challenges related to the collection of human-annotated data. 
Although RLHF does not require massive amounts of data to enhance performance, sourcing high-quality human preference data remains an expensive process~\cite{min2023factscore}.

To address these challenges, prior works have aimed to conduct preference optimization without relying on human preferences~\cite{paulus2018a, tian2024finetuning, wei2024measuring, roit-etal-2023-factually}. 
Such methods often require external metrics or complex filtering processes to establish preference pairs. 
For instance, \citet{paulus2018a} utilized lexical overlap (ROUGE) to assess salience and an entailment score to evaluate factual consistency. 
Similarly, \citet{tian2024finetuning} employed FactScore~\cite{min2023factscore} to gauge reward signals between generated summaries. 
However, as stated by Goodhart’s Law—‘\textit{When a measure becomes a target, it ceases to be a good measure}’—relying excessively on these imperfect metrics carries the risk of overfitting to the metrics alone~\cite{Strathern_1997, ramamurthy2023is}.

In response, we propose \textit{Model-based Preference Optimization} (MPO), a novel and straightforward approach that leverages the model's inherent summarization capabilities without relying on any human feedback or external metrics.
This method generates faithful summaries by aligning preferences between responses generated using different decoding strategies.
In particular, we utilize (1) a deterministic decoding strategy (\eg beam search decoding) to generate chosen samples and (2) a stochastic decoding strategy (\eg temperature sampling) to generate rejected samples. 
Therefore, our approach does not require any external knowledge or metrics to construct preference pairs.

Previous studies have shown that deterministic decoding strategies tend to generate outputs that are less surprising and closely aligned with the source text, while stochastic decoding introduces randomness, making it more prone to hallucinations~\cite{yang-etal-2018-breaking, welleck-etal-2020-consistency, holtzman2020curious, NEURIPS2022_df438caa}.
Specifically, \citet{wan-etal-2023-faithful-generation} presented empirical evidence indicating that beam search yields the most faithful summaries, while the randomness introduced by sampling reduces faithfulness.
Based on these findings, we align our model’s preference toward summaries generated via beam search rather than those randomly sampled.
As illustrated in Figure \ref{fig:radar_chart}, our approach outperforms models trained with standard supervised fine-tuning (SFT) or those optimized with human preferences (\eg PPO, DPO) in terms of faithfulness and relevance to the source text.

Our main contribution is Model-based Preference Optimization (MPO), a simple and straightforward approach for fine-tuning language models to improve abstractive summarization without relying on any human feedback or external metrics.
Our experimental results demonstrate that MPO outperforms models optimized with human preferences, offering superior overall performance and greater generalizability across diverse language models and datasets.

\begin{figure*}[t!]
    \centering
    \includegraphics[width=\textwidth]{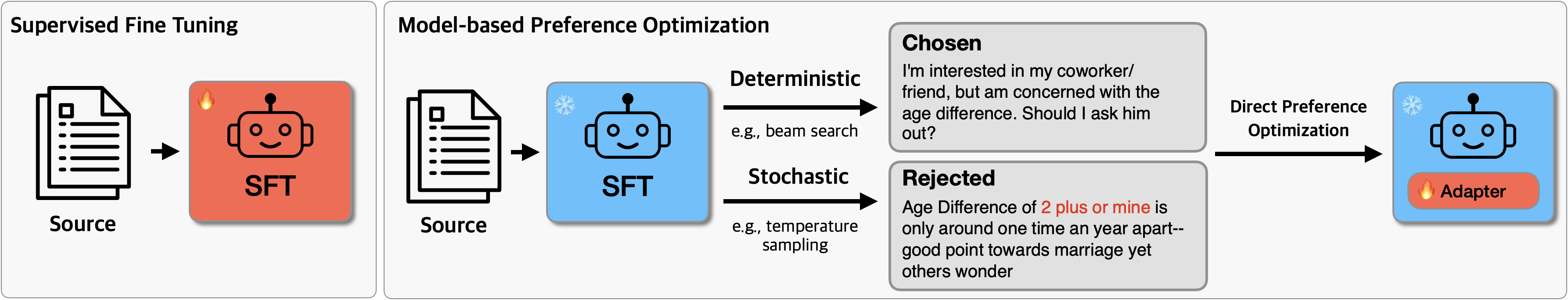}
    \caption{\textbf{Model-based Preference Optimization.} Our method follows a two-step process: 1) \textit{Supervised Fine-Tuning} (SFT): we fine-tune a pre-trained model (\ie LLM) on a given dataset. 2) \textit{Model-based Preference Optimization} (MPO): we build a preference dataset using different decoding strategies. In this step, the chosen samples are derived from deterministic decoding results, while the rejected samples utilize results generated by stochastic decoding.}
    \label{fig:overview}
\end{figure*}

\section{Preliminaries}
\label{sec:preliminaries}

\subsection{Problem Setup}

Let $\mathcal{V}$ denote the vocabulary for both input and output. We represent the input document as $\mathbf{x} \in \mathcal{X}$ and the output summary as $\mathbf{y} = \langle y_0, \ldots, y_T \rangle \in \mathcal{Y}$. The sequence $\mathbf{y}$ consists of $T+1$ elements, starting with the beginning-of-sequence token $y_0$ and ends with the end-of-sequence token $y_T$.

A language model (LM) is an auto-regressive model of a sequence distribution $P(\yy \!\mid\! \xx)$, where each conditional probability is parameterized by a neural network $p_{\theta}$. We assume that the model computes the probability of the entire generated text $\mathbf{y}$ using a common left-to-right decomposition. Thus, the distribution can be expressed as a product of conditional probabilities: 
\begin{equation*}
P(\mathbf{y} | \mathbf{x}) = \prod^{T}_{t=1}{p_{\theta}(y_t | \mathbf{y}_{<t}, \mathbf{x})}.  \end{equation*}

\subsection{LM for Summarization}

Given an input document $\xx$, the optimal summary $\mathbf{y}$ from the set of valid strings $\mathcal{Y}$ is obtained using a scoring function:
\begin{equation*}
    \mathbf{y}^{*} = \argmax_{\mathbf{y} \in \mathcal{Y}}
    {p_{\theta}(\mathbf{y} | \mathbf{x})}.
\end{equation*}
However, finding the optimal summary is not tractable. Therefore, the scoring function for the optimal string $\yy$ varies according to decoding strategies to approximate the best possible output. There are two types of decoding strategies: stochastic and deterministic.

\paragraph{Stochastic Decoding}

The simplest approach in decoding strategies is to sample directly from the probabilities predicted by the model. This method involves sampling from the conditional probability distribution at each step, represented as: 
\begin{equation*}
y_{\text{temp}} \sim P(y_t | \mathbf{x}, \mathbf{y}_{<t}).
\end{equation*}
However, this method exhibits high variance. To adjust for this variance, the temperature of the softmax function can be modified:
\begin{equation*}
   P(y_t | \mathbf{x}, \mathbf{y}_{<t}) = \mathrm{softmax}\left(\frac{p_{\theta}(y_t|\mathbf{x}, \mathbf{y}_{<t})}{\tau}\right),
\end{equation*}
where $\tau$ is the temperature parameter. 
Increasing $\tau$ causes the model’s conditional probability distribution to approach a uniform distribution, which can lead to the generation of random tokens that are irrelevant to the source documents.
Consequently, this increases the risk of the model producing hallucinations. 
For this reason, we classify samples generated through stochastic decoding as rejected samples in our preference dataset.

\paragraph{Deterministic Decoding} 

The other strategies are deterministic decoding algorithms. The most straightforward algorithm, called greedy decoding, simply selects the most probable token at each step~\cite{welleck-etal-2020-consistency}. This can be expressed as:
\begin{equation*}
y_{\text{greedy}} = \argmax_{y \in \mathcal{V}} \log p_{\theta}(y_t | \mathbf{y}_{<t}, \mathbf{x}).
\end{equation*}

In contrast to greedy decoding, beam search decoding considers the top-$k$ candidates for token generation. At each time step $t$, it tracks the $k$ most likely sequence hypotheses, where $k$ is the beam size. The output sequence can be represented as: 
\begin{equation*}
    \yy_{\text{beam}} = \argmax_{y \in \mathcal{V}} \sum_{t=1}^{L}{ \log p_{\theta}(y_t | \mathbf{y}_{<t}, \mathbf{x})},
\end{equation*}
where $L$ is the length of the final candidate sequence. 
These deterministic decoding strategies tend to produce tokens that are more closely related to the source document, resulting in more faithful summaries than those generated by stochastic decoding strategies. Therefore, we align our model’s preference toward summaries generated via the deterministic decoding strategies and define them as chosen samples in our preference dataset.

\section{Proposed Method}
\label{sec:method}

In this section, we detail our process for encouraging faithfulness in abstractive summarization. We follow the typical pipelines of preference optimization~\cite{dpo, ziegler2020finetuning, NEURIPS2020_1f89885d, NEURIPS2022_b1efde53}. However, by leveraging the differences between deterministic and stochastic decoding strategies, our pipeline does not require any external knowledge (\eg evaluation metrics) or human feedback. This pipeline is depicted in Figure~~\ref{fig:overview}.

\subsection{Supervised Fine-Tuning (SFT)}

For the summarization task, we first fine-tune a pre-trained language model using supervised learning on training data (\ie ground truth data), denoted as $\mathcal{D}^{\text{train}} = \{ (\mathbf{x}, \mathbf{y}_{\text{ref}}) \}$. 
Based on this supervised fine-tuning (SFT) approach, the model is trained to generate a single-sentence summary from a source document. In this work, we utilize existing SFT models with minimal modifications or apply SFT to pre-trained language models using QLoRA~\cite{dettmers2023qlora}.

\subsection{Preference Optimization}

For preference optimization, we employ Direct Preference Optimization (DPO,~\citeauthor{dpo},~\citeyear{dpo}). DPO simplifies the process by eliminating the need for an explicit reward function, making it preferable to RL-based algorithms, which incur significant computational costs by training multiple language models and sampling from the policy.

Given a dataset of preference pairs $\mathcal{D} = \{ (\xx_{i}, \yy^w_{i}, \yy^l_{i}) \}_{i=1}^N$, where $\xx_{i}$ represents source documents, $\yy^w_{i}$ are chosen responses, and $\yy^l_{i}$ are rejected responses, the probability of observing a preference pair is modeled using the Bradley-Terry model~\cite{bradley1952rank}:
\begin{equation*}
    p(\yy^w \succ \yy^l) = \sigma(r(\xx, \yy^w) - r(\xx, \yy^l)),
\end{equation*}
where $\sigma$ is the sigmoid function, and $r(\cdot, \cdot)$ is a reward function. 

\citet{dpo} demonstrated that models directly learn this policy from collected data without modeling the reward function. In other words, the 2-stage policy can be simplified into 1-stage policy. DPO loss can be expressed as:
\begin{equation*}
\begin{aligned}
&\mathcal{L}_\text{DPO}(\pi_{\theta}; \pi_\text{ref}) = \\
&-\mathbb{E}_{(\xx, \yy^w, \yy^l)\sim \mathcal{D}} \bigg{[}\log \sigma \bigg{(}\beta \log \frac{\pi_{\theta}(\yy^w\mid \xx)}{\pi_\text{ref}(\yy^w\mid \xx)} \\
& \qquad \qquad \qquad \qquad \qquad - \beta \log \frac{\pi_{\theta}(\yy^l\mid \xx)}{\pi_\text{ref}(\yy^l\mid \xx)}\bigg{)}\bigg{]},
\end{aligned}
\end{equation*}
where $\pi_{\text{ref}}$ is the SFT model and $\beta$ is a coefficient that controls the trade-off between reward and divergence. By optimizing this objective, the model aligns with the reward function while remaining close to the pre-trained reference model, thus minimizing over-optimization~\cite{tian2024finetuning}.

\subsection{Constructing Preferences Pairs without Human Feedback}

By exploiting the differences between deterministic and stochastic strategies, we construct a dataset of preference pairs, denoted as \(\mathcal{D}^{\text{valid}} = \{ (\mathbf{x}, \mathbf{y}^{w}_{\text{beam}}, \mathbf{y}^{l}_{\text{temp}}) \}\). This strategy is based on the observation that deterministic decoding typically produces more factual summaries~\cite{wan-etal-2023-faithful-generation}. This significant difference in output quality suggests that summaries generated through beam search decoding can be used as chosen samples, while those from temperature sampling can be designated as rejected samples. We then conduct preference optimization with this generated data to refine the language model, ensuring it avoids generating hallucinated or irrelevant text.

\begin{table*}[t]
\resizebox{1\textwidth}{!}{%
\begin{tabular}{c|lcccccc}
\noalign{\hrule height 1pt} 
\multirow{2}{*}{\small \textbf{\begin{tabular}[c]{@{}c@{}}Dataset\\ (Model)\end{tabular}}} & \multicolumn{1}{c|}{\multirow{2}{*}{\small\textbf{Method}}} & \multicolumn{1}{c|}{\multirow{2}{*}{\small\textbf{\begin{tabular}[c]{@{}c@{}}Response\\Ratio\end{tabular}}}} & \multicolumn{2}{c|}{\small\textit{\textbf{Faithfulness}}} & \multicolumn{2}{c|}{\small\textit{\textbf{Relevance}}} & \small\textit{\textbf{Similarity}} \\ \cline{4-8} 
 & \multicolumn{1}{c|}{} & \multicolumn{1}{c|}{} & \small\textbf{AlignScore $(\uparrow)$} & \multicolumn{1}{c|}{\small\textbf{FactCC $(\uparrow)$}} & \small\textbf{BARTScore $(\uparrow)$} & \multicolumn{1}{c|}{\small\textbf{BS-FACT $(\uparrow)$}} & \small\textbf{ROUGE-L $(\uparrow)$} \\
\hline
\hline
\multirow{9}{*}{\begin{tabular}[c]{@{}c@{}}TL;DR\\ (GPT-J)\end{tabular}} & \multicolumn{7}{l}{\cellcolor[HTML]{EFEFEF}\footnotesize\textit{with ground-truth data}} \\ \cline{2-8} 
 & \multicolumn{1}{l|}{SFT} & \multicolumn{1}{c|}{81.2\% (99.4\%)} & 89.21 (83.54) & \multicolumn{1}{c|}{64.18 (53.48)} & -1.25 (-1.63) & \multicolumn{1}{c|}{91.53 (90.30)} & 26.74 (26.01) \\
 & \multicolumn{1}{l|}{SFT++} & \multicolumn{1}{c|}{93.8\% (99.7\%)} & 87.29 (82.30) & \multicolumn{1}{c|}{61.50 (57.05)} & -1.37 (-1.63) & \multicolumn{1}{c|}{91.06 (90.11)} & \textbf{27.47 (26.53)} \\ \cline{2-8} 
 & \multicolumn{7}{l}{\cellcolor[HTML]{EFEFEF}\footnotesize\textit{with human feedback (preference dataset)}} \\ \cline{2-8} 
 & \multicolumn{1}{l|}{PPO} & \multicolumn{1}{c|}{100.0\% (100.0\%)} & 83.10 (75.88) & \multicolumn{1}{c|}{54.40 (47.52)} & -1.35 (-1.80) & \multicolumn{1}{c|}{91.32 (89.78)} & 23.55 (23.28) \\
 & \multicolumn{1}{l|}{DPO} & \multicolumn{1}{c|}{98.3 (99.8\%)} & 88.12 (82.55) & \multicolumn{1}{c|}{61.70 (54.09)} & -1.33 (-1.65) & \multicolumn{1}{c|}{91.27 (90.22)} & 27.24 (26.28) \\ 
    & \multicolumn{7}{l}{{\cellcolor[HTML]{EFEFEF}\footnotesize\textit{without human feedback}}} \\ 
 \cline{2-8} 
 & \multicolumn{1}{l|}{Preferred-FT} & \multicolumn{1}{c|}{66.8\% (99.6\%)} & 89.90 (82.04) & \multicolumn{1}{c|}{\textbf{76.58 (64.48)}} & -1.39 (-1.73) & \multicolumn{1}{c|}{91.24 (90.09)} & 24.38 (24.39) \\
 & \multicolumn{1}{l|}{MPO (Ours)} & \multicolumn{1}{c|}{99.9\% (99.9\%)} & \textbf{91.61$^{*}$ (86.82$^{*}$)} & \multicolumn{1}{c|}{72.10$^{*}$ (59.39$^{*}$)} & \textbf{-1.10$^{*}$ (-1.41$^{*}$)} & \multicolumn{1}{c|}{\textbf{92.20$^{*}$ (91.20$^{*}$)}} & 26.10 (26.49) \\ 
 \noalign{\hrule height 1pt}
\end{tabular}
}
\caption{\textbf{Results of the GPT-J model on the TL;DR dataset.} We compared our Model-based Preference Optimization (MPO) with two main baselines: \textit{supervised fine-tuning} and \textit{human preference}. All main results are based on a beam search decoding strategy, while the results in parentheses are based on a greedy decoding strategy. MPO showed overall better performance in terms of \textit{faithfulness} and \textit{source relevance} compared to other baselines. The SFT model is a fine-tuned model on the training split and the SFT++ model is the SFT model further fine-tuned on the validation split. PPO and DPO are SFT models optimized on human-preference datasets. Preferred-FT is a model fine-tuned only on the chosen samples of MPO. $*$ indicates statistical significance (p-value $< 0.001$) based on a T-test compared to DPO.}
\label{table:main_small}
\end{table*}

\renewcommand{\thefootnote}{\arabic{footnote}}

\section{Experiments}
\label{sec:setup}

\subsection{Experimental Setup}
\paragraph{Dataset}
We used the TL;DR dataset and the eXtreme Summarization (XSUM) dataset \cite{tldr, xsum}. The TL;DR dataset is constructed by Reddit posts and their corresponding TL;DR summaries, while the XSUM dataset consists of BBC articles and their single-sentence summaries. Both datasets are widely used for abstractive summarization tasks.

\paragraph{Models}
We utilized GPT-J (6B)~\cite{gpt-j}, Mistral-7B~\cite{mistral} and LLaMA2-7B~\cite{touvron2023llama}. For GPT-J model, we used a checkpoint from Huggingface\footnotemark[1] that was already fully fine-tuned on the train dataset. For LLaMA2-7B and Mistral-7B models, we performed Supervised Fine-Tuning (SFT) on each training dataset using QLoRA, and then merged the adapter into the models for further preference optimization experiments. 

\footnotetext[1]{\texttt{CarperAI/openai\_summarize\_tldr\_sft}}

\paragraph{Evaluation Metrics}
\label{subsec:metric}

We adopt the evaluation protocol proposed by \citet{chae-etal-2024-mitigating}. They categorized the evaluation into three key divisions: \textit{Faithfulness}, \textit{Relevance} (with the source), and \textit{Similarity} (with the target). For \textit{Faithfulness}, we used AlignScore \cite{zha2023alignscore} and FactCC \cite{kryscinski-etal-2020-evaluating}. To measure \textit{Relevance}, we employed BARTScore \cite{yuan2021bartscore} and BS-FACT. Lastly, to evaluate \textit{Similarity}, we used ROUGE-L.

\paragraph{Implementation Details}
For the SFT training, we utilized QLoRA with a batch size of 2 and a learning rate of 1e-4, training for one epoch in training split. After training, the SFT-trained QLoRA was merged with the pre-trained model. 
For preference optimization, we set the DPO hyperparameter β to 0.5. The learning rate was set to 1e-4 with a batch size of 4, and training was conducted for one epoch on the validation split. During summary generation, the maximum number of generated tokens was limited to 50. For beam search decoding, we used beam size of 6. For temperature sampling, we employed temperatures of 5.0 for GPT-J, and 1.0 for Mistral-7B and LLaMA2-7B.

\paragraph{Baselines}
We compared our method with two main baselines: \textit{supervised fine-tuned models} and \textit{human preference optimized models}. First, we compared our approach to models fine-tuned using ground-truth data or summaries generated via deterministic decoding. Second, we compared our method to PPO and DPO models trained on human preference pairs to demonstrate that the contrast between beam search decoding and random sampling is more effective than human-annotated preferences in terms of faithfulness.

\textbf{SFT} is a fine-tuned model on the train split of each dataset. \textbf{SFT++} is a model further trained on a validation split from the SFT model. \textbf{Preferred-FT} is fine-tuned to maximize likelihood only on the chosen samples (\ie $\yy_{\text{beam}}$). \textbf{PPO} and \textbf{DPO} are optimized from SFT models on human preference dataset provided by \citet{NEURIPS2020_1f89885d}. For PPO, we used a Huggingface checkpoint\footnotemark[2], already optimized with the provided human preference dataset. For DPO, we optimized in the same way as MPO but with the human preference dataset.

\footnotetext[2]{\texttt{CarperAI/openai\_summarize\_tldr\_ppo}}


\begin{table}[t]
\resizebox{1.0\columnwidth}{!}{%
\begin{tabular}{c|c|l|ccc}
\noalign{\hrule height 1pt}
\small\textbf{Dataset}& \small\textbf{Model} & \small\textbf{Method} & \small\textbf{AlignScore $(\uparrow)$}  & \small\textbf{BARTScore $(\uparrow)$}  & \small\textbf{ROUGE-L $(\uparrow)$} \\
\hline
\hline
\multirow{12.5}{*}{\rotatebox{90}{TL;DR}} & \multirow{4}{*}{GPT-J} & SFT   &  89.21 (83.54) &   -1.25 (-1.63)    & 26.74 (26.01)   \\
&& SFT++   &  87.29 (82.30) &   -1.37 (-1.63)    & \textbf{27.47 (26.53)}   \\
&& Preferred-FT   &  89.90 (82.04) &   -1.39 (-1.73)    & 24.38 (24.39)   \\
&& MPO (Ours)   &  \textbf{91.61 (86.82)} &   \textbf{-1.10 (-1.41)}    & 26.10 (26.49)   \\
\cline{2-6}   
& \multirow{4}{*}{Mistral} & SFT    &    87.85 (82.74)    & -1.48 (-1.81)  &  \textbf{25.32 (25.02)}  \\
&& SFT++   &  86.66 (82.10) &   -1.44 (-1.83)    & 25.27 (24.66)   \\
&& Preferred-FT   &   83.96 (79.70) &   -1.63 (-1.82)    & 22.57 (22.23)   \\
&& MPO (Ours)  &  \textbf{92.12 (89.39)} &   \textbf{-1.25 (-1.37)}    &  24.85 (25.01)   \\
\cline{2-6}   
& \multirow{4}{*}{LLaMA2}  & SFT    &   84.92 (77.68)   & -1.65 (-2.05)  &  24.31 \textbf{(23.33)}   \\
&& SFT++   &  \textbf{87.93} (78.03) &   \textbf{-1.41} (-2.05)    & \textbf{24.79} (22.89)   \\
&& Preferred-FT   &  81.10 \textbf{(79.58)} &   -1.74 \textbf{(-1.85)}    & 22.73 (22.47)   \\
&& MPO (Ours)  &  85.33 (78.03)       & -1.64 (-2.03)     &  24.16 (23.29)  \\
\hline                      
\multirow{12.5}{*}{\rotatebox{90}{XSUM}}  & \multirow{4}{*}{GPT-J} & SFT    &    64.01 (52.66)    & -1.59 (-1.97)  & 25.13 (24.41)   \\
&& SFT++   &  62.47 (49.91) &   -1.62 (-2.00)    & \textbf{25.58} (24.66)   \\
&& Preferred-FT   &  \textbf{66.42} (40.71) &   -1.68 (-2.13)    & 17.61 (20.21)   \\
&& MPO (Ours)  &  65.26 \textbf{(54.39)} &   \textbf{-1.58 (-1.95)}    & 25.25 \textbf{(24.72)}   \\
\cline{2-6}   
& \multirow{4}{*}{Mistral} & SFT    &   66.31 (60.00)   &   -1.96 (-1.97)    &   30.65 (31.16)  \\
&& SFT++   &  64.99 (60.17) &   \textbf{-1.74} (-1.96)    & 30.76 (30.72)   \\
&& Preferred-FT   &  63.74 (61.14) &   -2.53 (-3.31)    &  21.17 (18.57)   \\
&& MPO (Ours)  &  \textbf{68.58 (64.57)}      &  -1.85 \textbf{(-1.90)}    &  \textbf{31.11 (31.35)}   \\
\cline{2-6}
& \multirow{4}{*}{LLaMA2}  & SFT    &   65.80 (57.57)  &   -1.80 (-2.06)    &  \textbf{30.36} (27.76) \\
&& SFT++   &  67.20 (57.45) &   \textbf{-1.74} (-2.08)    & 29.23 (27.85)   \\
&& Preferred-FT   &  46.96 (39.86) &   -2.01 (-2.24)    & 24.36 (23.41)   \\
&& MPO (Ours)  &  \textbf{67.31 (60.48)}       &  -1.81 \textbf{(-2.02)}    &  30.32 \textbf{(28.36)}  \\
\noalign{\hrule height 1pt}
\end{tabular}
}
\caption{\textbf{Comparison of MPO with SFT}. MPO demonstrates generally robust results across various language models on both the TL;DR and XSUM datasets. The results are based on a beam search decoding strategy, while the results in parentheses are based on a greedy decoding strategy.}
\label{table:models}
\end{table}

\subsection{Comparison with Fine-Tuned Models}

In Table~\ref{table:main_small}, MPO consistently outperforms fine-tuned baselines (\ie SFT, SFT++, Preferred-FT). 
SFT++ and Preferred-FT did not significantly improve over SFT. 
However, MPO shows a substantial increase of up to 3.28 in AlignScore, 7.92 in FactCC, 0.22 in BARTScore, and 0.9 in BS-FACT over SFT. 
These results suggest that our approach is more effective at mitigating hallucinations than simply fine-tuning with either gold summaries or summaries generated through deterministic decoding.

In Table~\ref{table:models}, MPO demonstrates robust and generally applicable results across various language models on both the TL;DR and XSUM datasets. MPO generally exhibits a lower ROUGE-L score compared to SFT and SFT++. ROUGE-L measures the lexical similarity between generated summaries and the target summaries (\ie reference or gold) rather than the source document. However, higher lexical similarity does not necessarily imply better summary quality \cite{king2022dont, schluter-2017-limits, ng-abrecht-2015-better}. Thus, while ROUGE-L is a widely used metric, it is not the primary focus of our evaluation.

\begin{table}[t!]
\centering
\resizebox{1.0\columnwidth}{!}{%
\begin{tabular}{c|cc|cc}
\noalign{\hrule height 1pt}
\multirow{2}{*}{\textbf{GPT-3.5}} & \multicolumn{2}{c|}{\textbf{SFT (vs. MPO)}} & \multicolumn{2}{c}{\textbf{DPO (vs. MPO)}} \\ \cline{2-5} 
 & \textbf{Greedy} & \textbf{Beam} & \textbf{Greedy} & \textbf{Beam} \\ \cline{2-5} 
 \hline\hline
\# of compared samples & 6061 & 5376 & 5962 & 5332 \\
MPO win rate (\%) & \textbf{51.30} & \textbf{59.36} & \textbf{50.27} & 47.30 \\ \noalign{\hrule height 1pt}
\end{tabular}
}
\caption{\textbf{GPT-3.5 win rates on TL;DR dataset.} 
MPO generally outperformed the SFT, while the results varied when compared to DPO depending on the decoding method. Samples where both models produced identical outputs were excluded from the win rate calculation.}
\label{table:win_rate}
\end{table}
\begin{table}[t]
\centering
\resizebox{1.0\columnwidth}{!}{%
\begin{tabular}{p{1.5cm}p{6.5cm}}
\toprule
\small\textbf{Method} & \small\textbf{Text} \\ 
\midrule

Source & { 
TITLE: [19/f] What does this guy [20/m] actually want from me? POST: $\ldots$
became really good friends, $\ldots$ We then somehow from kissing gently $\ldots$
basically said he likes me but nothing can happen because I'm not his type... I JUST DON'T KNOW WHAT THE BOY WANTS FROM ME}. \\
\midrule
\normalsize{SFT} & {ive been friends with a guy for a while, then we kissed, \textbf{\redc{then we didn't, then we did again, then we didn't, then we did again.}}} \\
\midrule
\normalsize{DPO } & I don't know what the boy wants from me, \textbf{\redc{and I don't know what I want from the boy.}} \\
\midrule
\normalsize{MPO (Ours)} & Became really good friends with a guy, then we kissed, then he said he likes me but I'm not his type. What does he want from me? \\
\bottomrule
\end{tabular}%
}
\caption{\textbf{Example summaries of MPO model and human preference optimized model.} Inconsistent words are highlighted in \textbf{\redc{red}}. The summary generated by the MPO model is clearly superior to those by SFT and DPO (w/ human pref.) models in terms of faithfulness and source relevance.}
\label{table:compare_dpo}
\end{table}

\subsection{Comparison with Human Preference Optimized Models}

In Table \ref{table:main_small} and \ref{table:win_rate}, we compared MPO with human preference optimized models (\eg PPO, DPO). Based on the automatic metrics in Table \ref{table:main_small},  MPO consistently outperforms the human preference optimized models.
As noted in \citet{hosking2024human}, using human preference datasets can sometimes underestimate the aspect of faithfulness. 

On the other hand, as shown in Table \ref{table:win_rate}, MPO did not demonstrate dominant performance in the win rate evaluation based on GPT-3.5. For details on the win rate prompts, refer to Appendix \ref{appendix:win_rate_prompt}.
This discrepancy arises because summary evaluation involves multiple factors~\cite{hosking2024human, yuan2021bartscore}. 
While MPO excels in faithfulness and source relevance,  it may fall short in areas such as fluency (refer to Table \ref{table:compare_dpo}). 
Furthermore, human preference optimized models were trained on significantly more data pairs, utilizing multiple pairs per source text, whereas MPO was optimized using only one pair per source.


\begin{table}[t!]
\centering
\resizebox{0.9\columnwidth}{!}{%
\begin{tabular}{c|c|c}
\toprule
\textbf{Group} & \textbf{Selected} & \textbf{\# Samples} \\ 
\hline\hline
 \multirow{2}{*}{Group A (MPO wins)} & MPO & 35 \\ 
                    & DPO & 15 \\ 
\midrule
 \multirow{2}{*}{Group B (DPO wins)}  & MPO & 16 \\
                    & DPO & 34 \\
\bottomrule
\end{tabular}
}
\caption{\textbf{Results of human evaluation.} MPO achieves an overall win rate of 51\% compared to the DPO.}
\label{table:human_win_rate}
\end{table}

\paragraph{Human Evaluation} 
To assess whether the automatic score (\ie AlignScore) aligns with human preference, we conducted human evaluations on 100 samples from the TL;DR dataset across two groups. More details are provided in Appendix \ref{appendix:human_eval_details}.

\begin{itemize}
    \item Group A: AlignScore of DPO ≤ 0.5 and AlignScore of MPO > 0.5
    \item Group B: AlignScore of DPO > 0.5 and AlignScore of MPO ≤ 0.5
\end{itemize}

In Table~\ref{table:human_win_rate}, MPO achieves an overall win rate of 51\% when combining results from Groups A and B. Notably, 70\% of MPO's summaries in Group A were evaluated superior, while only 32\% received favorable judgments in Group B. These results suggest that AlignScore aligns with human judgment to some extent, indicating that our evaluation method can yield results comparable to human evaluation.

\subsection{Comparison with Decoding Strategies}
Table \ref{table:decoding} shows the results of applying MPO models to various decoding strategies using the LLaMA2-7B model. Despite not being specifically optimized for various decoding strategies (\ie Nucleus \cite{holtzman2020curious}, ITI \cite{li2023inferencetime}, DoLa \cite{chuang2023dola}), MPO models are generally applicable to all decoding strategies and consistently produces enhanced summarization results compared to the standard SFT model in terms of faithfulness and relevance.

\section{Analysis}

\begin{table}[t!]
\resizebox{1\columnwidth}{!}{%
\begin{tabular}{c|c|ccc}
\noalign{\hrule height 1pt}

\begin{tabular}[c]{@{}c@{}}\small \textbf{Decoding}\\ \small 
 \textbf{Strategy}\end{tabular} & \textbf{Method} & \small{\textbf{AlignScore $(\uparrow)$}}  & \small{\textbf{BARTScore $(\uparrow)$}}  & \small{\textbf{ROUGE-L $(\uparrow)$}} \\
\hline
\hline
\multirow{2}{*}{Greedy} & SFT   & 77.68 & -2.05 & 23.33 \\
         & MPO & 78.03      & -2.03     & 23.29   \\
\hline
\multirow{2}{*}{Nucleus}& SFT  & 76.25 & -2.11 & 22.82 \\ 
      & MPO &    76.99   &  -2.09    &  22.79  \\      
\hline
\multirow{2}{*}{ITI}& SFT      & 76.95 & -1.88 & 23.15 \\      
       & MPO & 77.15 & -1.87 & 23.23 \\      
\hline
\multirow{2}{*}{DoLa} & SFT    & 82.47 & -1.76 & 24.61 \\
       & MPO           &      82.57      &    -1.75       &   24.55      \\
\hline
\multirow{2}{*}{Beam} & SFT     & 84.92 & -1.65 & 24.31 \\            
       & MPO & 85.33      & -1.64     & 24.16   \\            
\noalign{\hrule height 1pt}
\end{tabular}
}
\caption{\textbf{Results of applying various decoding strategies}. MPO aligns well with different decoding strategies. When combined with faithfulness-aware decoding strategies (\ie ITI, DoLA), it can lead to further improvements. The results are from using the LLaMA2-7B on the TL;DR dataset.}
\label{table:decoding}
\end{table}

\begin{table}[t]
\resizebox{1.0\columnwidth}{!}{%
\begin{tabular}{c|c|c|ccc}
\noalign{\hrule height 1pt}
\small\textbf{Dataset} & \small\textbf{Model} & \small\textbf{Method} & \small\textbf{AlignScore $(\uparrow)$} & \small\textbf{BARTScore $(\uparrow)$} & \small\textbf{ROUGE-L $(\uparrow)$} \\
\hline
\hline
\multirow{9.5}{*}{\rotatebox{90}{TL;DR}} & \multirow{3}{*}{GPT-J} & Beam search      &   \textbf{89.19}  & \textbf{-1.24}  & \textbf{27.00} \\
&& Sampling (temp1)   & 57.68  & -2.94  & 19.34 \\
&& Sampling (temp5)   & 24.66  & -6.89  & 8.73  \\
\cline{2-6}   
& \multirow{3}{*}{Mistral} & Beam search  & \textbf{87.47}  & \textbf{-1.46}  & \textbf{25.18} \\
&& Sampling (temp1)   & 58.70  & -3.14  & 18.43 \\
&& Sampling (temp5)   & 22.96  & -7.14  & 8.35  \\
\cline{2-6}   
& \multirow{3}{*}{LLaMA2}  & Beam search  & \textbf{84.72}  & \textbf{-1.65}  & \textbf{24.41} \\
&& Sampling (temp1)   & 64.23  & -2.71  & 19.69 \\
&& Sampling (temp5)   & 23.27  & -7.12  & 8.51  \\
\hline
\multirow{9.5}{*}{\rotatebox{90}{XSUM}} & \multirow{3}{*}{GPT-J} & Beam search & \textbf{64.55}  & \textbf{-1.59}  & \textbf{25.34} \\
&& Sampling (temp1)   & 28.12  & -2.99  & 17.77 \\
&& Sampling (temp5)   & 14.33  & -6.91  & 6.95  \\
\cline{2-6}   
& \multirow{3}{*}{Mistral}  & Beam search  & \textbf{66.76}  & \textbf{-1.96}  & \textbf{30.57} \\
&& Sampling (temp1)   & 43.48  & -2.81  & 22.81 \\
&& Sampling (temp5)   & 20.07  & -7.41  & 6.82  \\
\cline{2-6}   
& \multirow{3}{*}{LLaMA2}  & Beam search  & \textbf{66.57}  & \textbf{-1.81}  & \textbf{30.48} \\
&& Sampling (temp1)   & 47.65  & -2.49  & 23.76 \\
&& Sampling (temp5)   & 17.65  & -7.41  & 7.39  \\
\noalign{\hrule height 1pt}
\end{tabular}
}
\caption{\textbf{Evaluation reults of chosen and rejected samples}. Summaries generated with deterministic decoding (\eg beam search) outperformed those from stochastic decoding (\eg temperature-scaled sampling) across all metrics.}
\label{table:temp_small}
\end{table}

\subsection{Evaluation of Chosen and Rejected Samples}

Our key assumption is that deterministic generation yields summaries more relevant to the source document than stochastic generation for summarization tasks. In Table \ref{table:temp_small}, we compared the deterministic and stochastic generated summaries used in MPO training. The chosen samples consistently outperformed the rejected samples across all metrics. Our results align with the results of recent studies \cite{holtzman2020curious, wan-etal-2023-faithful-generation, NEURIPS2022_df438caa}.

However, these findings do not necessarily imply that deterministic generation is always less hallucinated than stochastic generation. Thus, we adjusted the temperature in stochastic sampling to encourage the generation of tokens that are unrelated to the source documents.

\subsection{Other Combinations for Preference Pairs}

\begin{table}[tb!]
\centering
\resizebox{\columnwidth}{!}{%
\begin{tabular}{c|ccccc}
\noalign{\hrule height 1pt}
\small\textbf{Combination}& \small\textbf{AlignScore $(\uparrow)$}  & \small\textbf{BARTScore $(\uparrow)$}  & 
\small\textbf{ROUGE-L $(\uparrow)$} \\  
\hline
\hline
\small{SFT} & \small{89.21}& \small{-1.25}& \small{26.74}\\
\small{$(\yy^{w}_{\text{beam}}, \yy^{l}_{\text{greedy}})$} &  \small{51.96}  & \small{-4.63} & \small{0.87}\\
\small{$(\yy^{w}_{\text{temp5}}, \yy^{l}_{\text{beam}})$} &  \small{87.59}   & \small{-1.36} & \small{27.24} & \\
\small{$(\yy^{w}_{\text{greedy}}, \yy^{l}_{\text{temp5}})$} &  \small{90.57}  & \small{-1.20}   & \small{26.87} \\
\small{$(\yy^{w}_{\text{beam}}, \yy^{l}_{\text{temp5}})$} &  \small{91.61}  & \small{-1.10}   & \small{26.10} \\
\noalign{\hrule height 1pt}
\end{tabular}
}
\caption{\textbf{MPO with different combinations of preference pairs.} The result show that using a deterministic decoding strategy pair significantly inhibit summarization ability. For pairs combining deterministic and stochastic decoding, setting beam search as the chosen and temperature-based sampling as the rejected maximizes the language model's summarization performance. The results are from using the GPT-J on the TL;DR dataset.}
\label{table:combinations}
\end{table}
\begin{table}[tb!]
\centering
\resizebox{\columnwidth}{!}{%
\begin{tabular}{c|ccccc}
\noalign{\hrule height 1pt}
\small\textbf{Pairs}& \small\textbf{ROUGE-1 $(\uparrow)$}  & \small\textbf{ROUGE-2 $(\uparrow)$}  & 
\small\textbf{ROUGE-L $(\uparrow)$} \\  
\hline
\hline
\small{$\yy^{w}_{\text{beam}} \;\text{vs.}\; \yy^{l}_{\text{greedy}}$} &  \small{47.38}  & \small{35.06} & \small{43.24}\\
\small{$\yy^{w}_{\text{greedy}} \;\text{vs.}\; \yy^{l}_{\text{temp5}}$} &  \small{12.93}  & \small{0.49}   & \small{9.00} \\
\small{$\yy^{w}_{\text{beam}} \;\text{vs.}\; \yy^{l}_{\text{temp5}}$} &  \small{10.56}  & \small{0.41}   & \small{7.40} \\
\noalign{\hrule height 1pt}
\end{tabular}
}
\caption{\textbf{ROUGE score comparison.} Deterministic decoding generated summaries exhibit high similarity, whereas there is low similarity between summaries generated by deterministic decoding and those generated by stochastic decoding.}
\label{table:rouge}
\end{table}

\paragraph{Deterministic Generation as Rejected Samples}

To assess whether improving the quality of rejected responses would enhance the model's summarization performance, we employed greedy decoding for the rejected responses. However, this approach resulted in a notable decline in summarization performance (see row 3 in Table \ref{table:combinations}). Examples of the generated samples are provided in Appendix \ref{appendix:example_cases}.

One reason for degradation is that the chosen and rejected samples are too similar, causing confusion for the model. In Table \ref{table:rouge}, we measured the similarity between the summaries produced by the two decoding methods. The summaries generated by beam search decoding and greedy decoding achieved very high ROUGE scores. This suggests that using overly similar summaries as chosen and rejected responses in preference optimization can have adverse effects \cite{pal2024smaug}.

\paragraph{Stochastic Generation as Chosen Samples}

In table \ref{table:combinations}, we instead used stochastic decoding for the chosen samples. While this approach did not result in severe degeneration, it reduced faithfulness compared to the original SFT model. This suggests that if the chosen samples have lower source alignment than the rejected ones, preference optimization can degrade the model's existing summarization performance in terms of faithfulness and relevance.

\begin{figure}[t]
      \centering
      \includegraphics[width=\columnwidth]{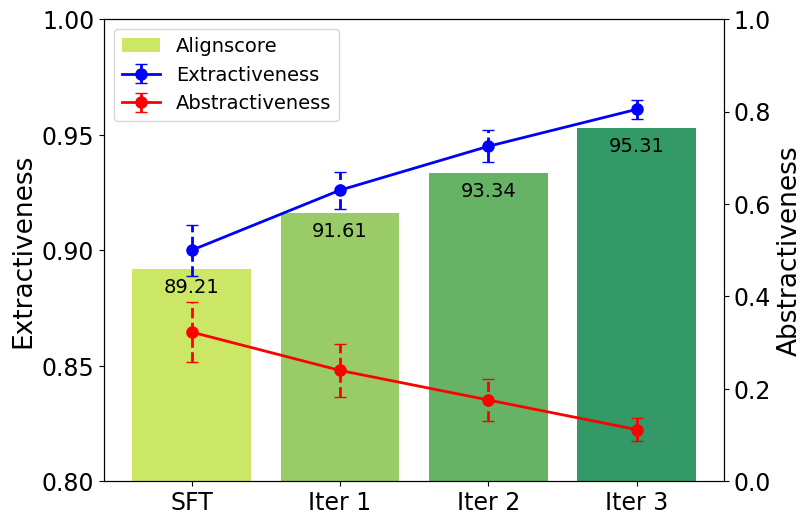}
    \caption{\textbf{Analysis for each training iteration.} The average abstractiveness of summaries generated for the TL;DR test set across training iterations, measured by the MINT score, with dotted lines indicating variance. The average extractiveness is measured by extractive fragment coverage.}
    \label{fig:iterative_results}
\end{figure}

\subsection{Faithfulness-Abstractiveness Tradeoff from Iterative Training}

Recent studies by \citet{irpo} and \citet{chen2024self} have shown that iteratively constructing the preference dataset using the trained model from the previous iteration improves dataset quality. Building on these works, our approach extends MPO to iterative MPO. In this experiment, we used summaries generated via beam search from the previous iteration as the chosen samples, while summaries generated through random sampling from the initial SFT model were used as rejected samples. To adapt to the continuous improvements in model performance, we dynamically adjusted the task difficulty by progressively lowering the temperature settings—5.0, 3.0, and 1.0—for each iteration.

We observed a notable trend where the model increasingly produced more extractive summaries, often directly incorporating sentences from the source documents. This trend can be attributed to the slightly extractive nature of the summaries generated by the SFT model using beam search decoding, which were used as the chosen samples~\cite{ladhak-etal-2022-faithful}. In other words, iterative MPO training may suppress the model’s creativity. Consequently, as shown in Figure \ref{fig:iterative_results}, the model's faithfulness improved with increased extractiveness over successive iterations\footnotemark[3].

\footnotetext[3]{To quantitatively assess the abstractiveness and extractiveness, we utilized the MINT~\cite{dreyer2023evaluating} and \textit{extractive fragment coverage}~\cite{grusky2018newsroom}, respectively.}

\paragraph{Qualitative study}

In Appendix~\ref{appendix:example_cases}, Table \ref{table:extractive_results} provides an example of summaries generated by the SFT model and by the MPO model at different iterations in response to a given prompt. As the iterations progress, the summaries tend to become more extractive for the document. Notably, the summary generated in the third iteration is quite similar to the title.

\subsection{Encoder-Decoder Model}

To verify the generalizability of our method across different model architectures, we evaluated it using an encoder-decoder model, such as BART~\cite{lewis2019bart}. As shown in Table \ref{table:bart}, MPO outperforms SFT in terms of AlignScore, improving from 61.86 to 66.42. These results demonstrate that our approach can be applied to encoder-decoder models as well. Additionally, we compared MPO with another decoding strategy baseline, \textit{Faithfulness-aware Lookahead}~\cite{wan-etal-2023-faithful-generation}, which has shown effectiveness with encoder-decoder models. Interestingly, by using summaries from Faithfulness-aware Lookahead as the chosen samples instead of the beam search summaries (\ie MPO*), MPO* increased the AlignScore by 2.43 over MPO. This suggests that incorporating more effective decoding strategies within MPO can further enhance summarization performance.


\begin{table}[t!]
\centering
\resizebox{1.0\columnwidth}{!}{%
\begin{tabular}{@{}l|cccc@{}}
\noalign{\hrule height 1pt}
\small\textbf{Method}& \small\textbf{AlignScore ($\uparrow$)}  & \small\textbf{BARTScore($\uparrow$)}  & 
\small\textbf{ROUGE-L($\uparrow$)} \\  
\hline\hline
 \small{SFT} & \small{61.86}  & \small{-1.80}   & \textbf{\small{36.42}} \\
 \small{MPO $(\yy^{w}_{\text{beam}}, \yy^{l}_{\text{temp}})$}  &\small{66.42}  & \small{-1.80}   & \small{35.78} \\
 \cline{1-5}
 \small{Lookahead~\cite{wan-etal-2023-faithful-generation}} &  \small{67.78}  & \small{-1.76} & \small{34.3}\\
 \small{MPO* $(\yy^{w}_{\text{Lookahead}}, \yy^{l}_{\text{temp}})$
} & \textbf{\small{68.85}}   & \textbf{\small{-1.73}} & \small{34.93} & \\
\noalign{\hrule height 1pt}
\end{tabular}
}
\caption{\textbf{Results of experiments for the encoder-decoder model on XSUM dataset.} MPO outperforms SFT in terms of factuality. The summarization performance of MPO can be further improved by using enhanced decoding strategy (\eg Lookahead) instead of beam search decoding}
\label{table:bart}
\end{table}

\section{Related Work}
\label{sec:related_work}

In the realm of auto-regressive language models, there are two primary approaches aimed to enhance the model's summarization capabilities: adjusting the learning algorithm or refining the decoding strategy~\cite{Welleck2020Neural}. The former involves updating the model's parameters through a learning objective, while the latter entails improving the decoding algorithm during generation while maintaining the existing pre-trained parameters frozen. In this paper, we will review two approaches in abstractive summarization aimed at alleviating hallucination.

\paragraph{Decoding Strategies}

Several methods have been proposed to rectify hallucinations during generation. Inference-time intervention (ITI) shifts activations along truth-correlated directions~\cite{li2023inferencetime}, repeating the same intervention auto-regressively until the entire answer is generated. Decoding by contrasting layers (DoLa) uses an early-exit strategy by contrasting the differences in logits obtained from projecting the later layers versus earlier layers~\cite{chuang2023dola}. Lastly, \citet{wan-etal-2023-faithful-generation} extend the idea of lookahead~\cite{lu-etal-2022-neurologic} to improve faithfulness in abstractive summarization, showing that the deterministic decoding strategy outperforms nucleus sampling~\cite{holtzman2020curious} in terms of faithfulness. However, it is important to note that decoding strategies do not change the underlying model.

\paragraph{Learning Algorithms}

To mitigate hallucinations, naively fine-tuning with faithfulness-aware objectives might seem straightforward. FactPegasus \cite{wan2022factpegasus} employs a tailored pre-training setup with contrastive learning to generate more faithful summaries. It modifies sentence selection by combining ROUGE and FactCC~\cite{kryscinski-etal-2020-evaluating}. However, this method risks overfitting to the metrics used, potentially degrading overall summarization performance~\cite{chae-etal-2024-mitigating}.

As an alternative, RL-based objectives can be utilized to enhance faithfulness~\cite{bohm-etal-2019-better, roit-etal-2023-factually, paulus2018a}. RL provides a natural path for optimizing non-differentiable objectives in LM-based generation. \citet{ramamurthy2023is} show that RL techniques generally align language models to human preferences better than supervised methods. On the other hand, Direct Preference Optimization (DPO)\cite{dpo} simplifies the process by eliminating the need for an explicit reward function of RL-based algorithms. Leveraging DPO, \citet{tian2024finetuning} have suggested optimizing language models for factuality in long-form text generation using FactScore~\cite{min2023factscore}.

In this paper, we train the underlying model to provide summaries faithful to source documents, based on findings from research on decoding strategies. Our approach does not require external metrics or human feedback during the optimization process. Furthermore, the model trained on our framework is versatile enough to integrate enhanced decoding techniques, thereby more effectively reducing hallucinations.

\section{Conclusion}
\label{sec:conclusion}

This study introduces Model-based Preference Optimization (MPO), a novel approach to improve the faithfulness and quality of abstractive summaries generated by Large Language Models (LLMs). Unlike traditional methods that rely heavily on costly human feedback, MPO leverages the model's inherent summarization capabilities to create a preference dataset using different decoding strategies. Our extensive experiments demonstrate that MPO significantly enhances the summarization performance, providing an efficient and scalable solution to address the challenges of hallucination in LLM-generated summaries.

\section*{Limitation}
\label{sec:limitation}
In our experiments, we employed QLoRA to maintain the performance of the SFT model. However, this method may have limited further performance improvements. The absence of comparative experiments leaves uncertainty about actual effectiveness of QLoRA. Additionally, due to constraints in our experimental environment, we limited experiments on 7B models, which raises concerns about the scalability of our approach.

During iterative training, we observed a trend where the model increasingly adopted an extractive approach, often replicating sentences from the input documents directly in the summaries. This trend poses a challenge to our goal of producing more faithful abstractive summaries.

\section*{Ethical Concerns}
We propose MPO, which leverages the outputs of a language model as a dataset for preference optimization, relying extensively on the outputs from the SFT model. Previous researches (\citet{sheng2019woman}, \citet{nangia2020crows}) has shown that self-supervised language models, which are trained on unlabeled web-scale datasets, can unintentionally learn and perpetuate social and ethical biases, including racism and sexism. If such biases are inherent within the data, our proposed self-feedback framework may unintentionally reinforce them. We used the TL;DR dataset for training, derived from Reddit posts, which may contain unmoderated and biased expressions. The presence of offensive content in this dataset risks influencing the model's outputs, potentially perpetuating these biases in further training within MPO. Moreover, as MPO progresses and the model increasingly favors extractive summarization, it may struggle to effectively paraphrase and filter out offensive expressions.

\section*{Acknowledgements} 

This research was supported by the National Research Foundation of Korea (NRF) grant (No. RS-2023-00222663, RS-2024-00345809, RS-2024-00333484, RS-2024-00414981) and the Institute of Information \& Communications Technology Planning \& Evaluation (IITP) grant (under the Leading Generative AI Human Resources Development, IITP-2024-RS-2024-00397085), both funded by the Korea government (MSIT).

\bibstyle{latex/acl_natbib.bst}
\bibliography{custom}

\appendix

\clearpage

\section{Appendix}
\label{sec:appendix}

\subsection{GPT-3.5 Judgment Prompts}
\label{appendix:win_rate_prompt}

\begin{figure}[h!]
      \centering
      \includegraphics[width=1.05\columnwidth]{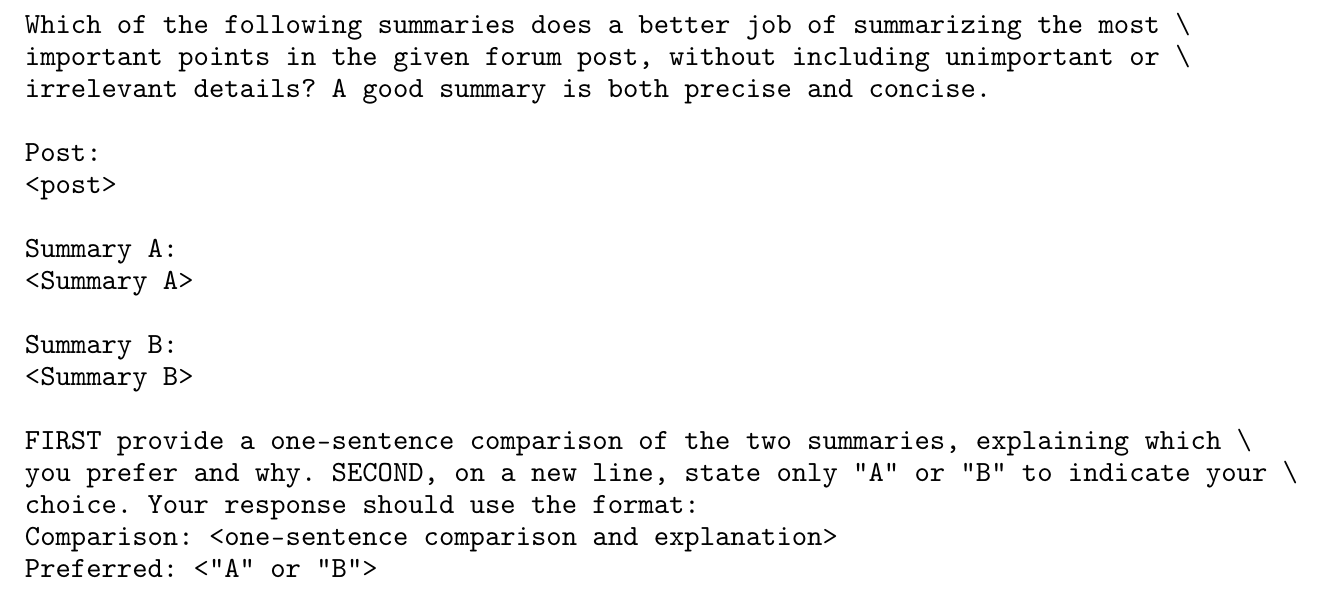}
    \caption{\textbf{Prompt of GPT-3.5 win rate.} }
    \label{fig:prompt}
\end{figure}

We used \textit{GPT-3.5-turbo} to evaluate win rates using prompts proposed in \citet{dpo}. The order of summaries or responses is randomly chosen for each evaluation. The prompt examples we used can be seen in Figure \ref{fig:prompt}.

\subsection{Human Evaluation Details}
\label{appendix:human_eval_details}

We sampled 100 instances from the TL;DR dataset for human evaluation. Instead of randomly sampling instances from the dataset, we selected instances to effectively assess the reliability of AlignScore \cite{zha2023alignscore} in comparison to human evaluation. Our goal was to determine if the automatic score (\ie AlignScore) aligns with human judgment. To achieve this, we divided the dataset into four groups:

\begin{itemize}
    \item Group A: AlignScore of DPO ≤ 0.5 and AlignScore of MPO > 0.5
    \item Group B: AlignScore of DPO > 0.5 and AlignScore of MPO ≤ 0.5
    \item Group C: AlignScore of DPO ≤ 0.5 and AlignScore of MPO ≤ 0.5
    \item Group D: AlignScore of DPO > 0.5 and AlignScore of MPO > 0.5
\end{itemize}

To ensure fairness and align with our primary goal, we evenly mixed Group A (MPO wins) and Group B (DPO wins) by sampling 50 instances from each group. We excluded instances from Group C and Group D because the differences between instances in those groups were minimal, making it challenging for human annotators to assess preferences based on just a few words.
    
\begin{figure}[h!]
      \centering
      \includegraphics[width=1.05\columnwidth]{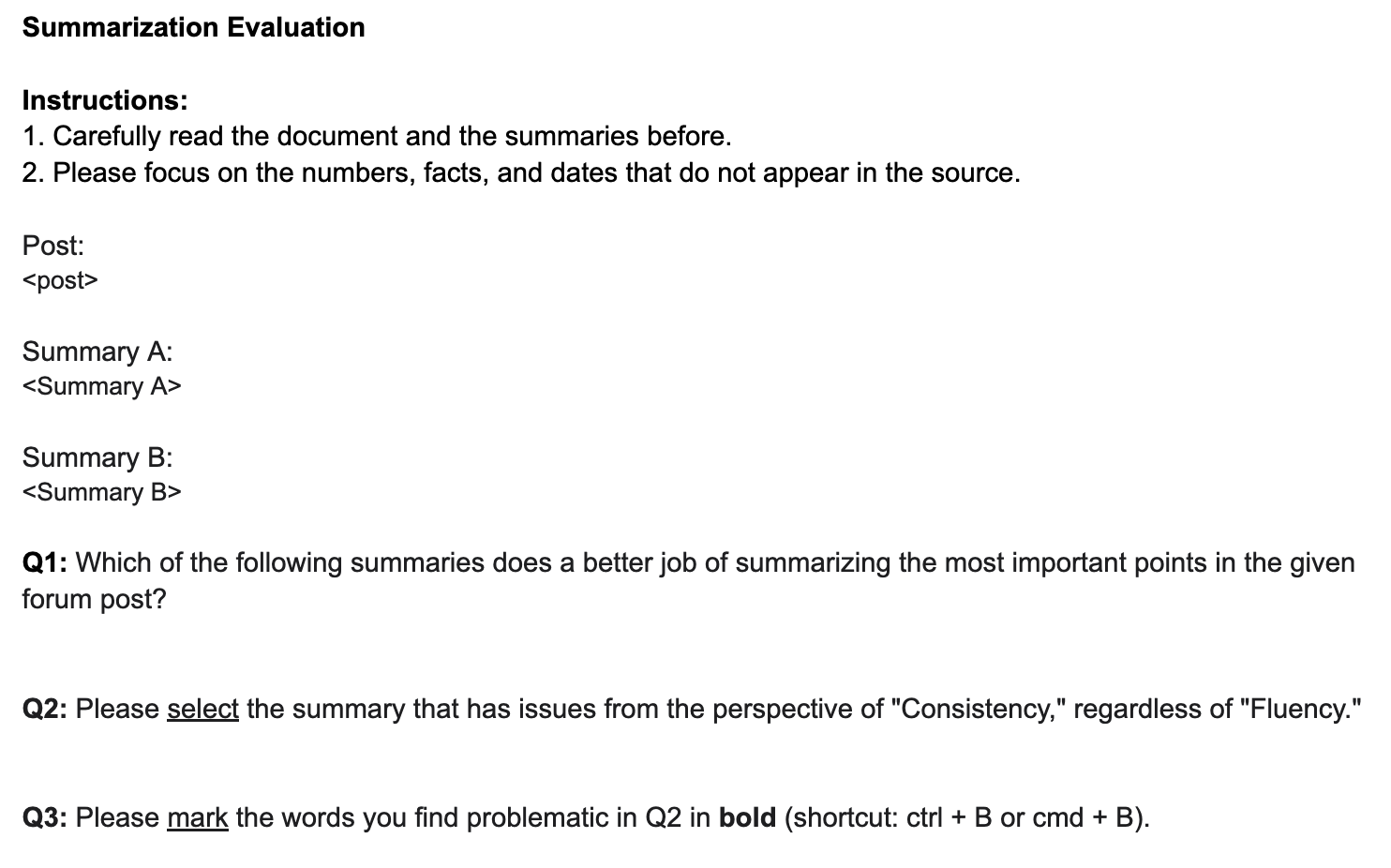}
    \caption{\textbf{Layout of human evaluation.} }
    \label{fig:layout_human_evaluation}
\end{figure}

We asked the annotators three questions. First, they were asked to choose the summary they considered better between the two provided summaries (Q1). Second, they were asked to select the summary with issues based on consistency with the source text (Q2). Finally, they were instructed to mark the parts of the selected summary they found problematic (Q3). For Q2, they could choose one of four responses: Summary A, Summary B, Neither, or Both. Figure \ref{fig:layout_human_evaluation} illustrates the layout format used in the survey.

\begin{table}[h!]
\centering
\resizebox{0.6\columnwidth}{!}{%
\begin{tabular}{c|c}
\toprule
\textbf{Selected} & \textbf{\# Samples} \\ 
\hline\hline
 
 MPO (ours)        & 29   \\ 
 DPO               & 36 \\ 
 Neither           & 10  \\
 Both              & 25 \\
\bottomrule
\end{tabular}
}
\caption{\textbf{Results of human evaluation on consistency.}}
\label{table:human_eval2}
\end{table}

Based on the responses to Q2 and Q3, we confirmed that our method produced summaries that were more faithful to the source text compared to DPO (see Table \ref{table:human_win_rate}). Interestingly, although both methods had similar win rates, DPO performed significantly worse in terms of consistency (see Table \ref{table:human_eval2}).

\paragraph{Participants}
We had 10 annotators, each of whom reviewed 10 samples, resulting in a total evaluation of 100 samples. All of raters were students of Seoul National University with a STEM focus. We appreciate the contributions of the following volunteers. The names are listed in random order:

Hyunbin Jin, Gihoon Kim, Minsoo Jo, Seunghyun Bae, Jewon Yeom, Seoyun Yang, Ijun Jang, Seul Lee, Junoh Park, Jinmyeong Choi

\subsection{Example Cases}
\label{appendix:example_cases}

Table \ref{table:degeneration} shows examples of summaries with different combinations of preference pairs. Table \ref{table:extractive_results} shows examples summaries from iterative preference optimization.

\begin{table*}[p]
\centering
\resizebox{0.95\textwidth}{!}{%
\begin{tabular}{p{2.7cm}p{12cm}}
\toprule
\small\textbf{Method} & \small\textbf{Text} \\ 
\midrule
\small{Source} & \small{SUBREDDIT: r/relationships TITLE: Is she [21] playing hard to get or uninterested in me?[22/M]POST: Hey guys first post here. So I'll try to make this quick, I've been out of the dating scene for a few years now and need advice with a girl i currently like. Her and I met in class and have been talking for not too long a month or so. We have tons in common I have gotten her number and we text every now and then (more on that later.) But I have really I've made the most progress in the past week. So everything explained further takes place in a span of a week. I've hung out with her a few times. The times we have hung out have been good we seem to hit it off. She's kinda touchy and takes lots of pictures of us (bff on sc if that means anything.) She said things like I'm special for getting to see her act in her goofy way. She even made and brought me a sandwich for when we were studying together. But ever since then she seems less interested in me and we do not text as often. Which is weird cuz that was to me at least the time we hit it off the most.  Before We had been texting all day but now barely at all and remember this is all in a span of less than a week. Most recently we were supposed to hangout but she said she forgot (which is a big red flag I know) but we did reschedule for another specific day. So is she uninterested? Playing hard to get? Or other? TL;DR:} \\
\small{MPO $(\yy^{w}_{\text{beam}}, \yy^{l}_{\text{greedy}})$} & \small{$\setminus$ufffd $\setminus$ufffd $\setminus$ufffd $\setminus$ufffd $\setminus$ufffd $\setminus$ufffd $\setminus$ufffd $\setminus$ufffd $\setminus$ufffd $\setminus$ufffd $\setminus$ufffd $\setminus$ufffd $\setminus$ufffd $\setminus$ufffd $\setminus$ufffd $\setminus$ufffd $\setminus$ufffd $\setminus$ufffd $\setminus$ufffd $\setminus$ufffd $\setminus$ufffd $\setminus$ufffd $\setminus$ufffd $\setminus$ufffd $\setminus$ufffd $\setminus$ufffd $\setminus$ufffd $\setminus$ufffd}
\\
\small{MPO $(\yy^{w}_{\text{beam}}, \yy^{l}_{\text{temp5}})$} & \small{Girl I like seems less interested in me after we hit it off for a week. Is she uninterested? Playing hard to get? Or other?} \\
\small{MPO $(\yy^{w}_{\text{temp5}}, \yy^{l}_{\text{beam}})$} & \small{ive been talking to a girl for less than a month and we seem to hit it off but now she seems less interested.} \\
\bottomrule
\end{tabular}%
}
\caption{\textbf{Example of summaries with different combinations of preference pairs.} In the case of $(\yy^{w}_{\text{beam}}, \yy^{l}_{\text{greedy}})$, the quality of the generated summaries significantly deteriorated. When there is a clear distinction between preferred data and rejected data, as observed in the two models below, the generated summaries remain similar even if the preferred and rejected data are swapped.}
\label{table:degeneration}
\end{table*}

\begin{table*}[t]
\centering
\resizebox{0.95\textwidth}{!}{%
\begin{tabular}{p{2.3cm}p{12cm}}
\toprule
\small\textbf{Method} & \small\textbf{Text} \\ 
\midrule
\small{Source} & \small{SUBREDDIT: r/AskReddit TITLE: \textbf{Any doctors or med students lurking? What is the process to become a doctor like?} POST: I would like to become an anesthesiologist one day. I have thought about this a lot and would like to hear what it's like from some insiders. I am really not the type who can sit behind a desk all day. I'm prepared to sit at desks for studying, but I can't do it for 30+ years. I don't want to wear a suit and tie. I like to be up running around. Nothing else really interests me aside from extreme sports. I want to put my work in now and come out making a high salary. I don't want to climb the rungs of a corporate ladder. I've had friends with dads in construction, working 80+ hours a week. My dad as an engineer and programmer worked 80+ hours a week for years and his salary still isn't as high as an anesthesiologists (it's great for an engineer). I really just want to become an anesthesiologist.TL;DR:} \\

\small{SFT} & \small{I want to become an anesthesiologist. I'm not sure if I'm going about it the right way.}\\

\small{MPO (iteration 1)} & \small{I want to become an anesthesiologist. I want to know what it's like from an insider. What is the process like?} \\

\small{MPO (iteration 2)} & \small{I want to become an anesthesiologist. I want to know what it's like from an insider. \textbf{What is the process to become a doctor like?}} \\

\small{MPO (iteration 3)} & \small{\textbf{Any doctors or med students lurking? What is the process to become a doctor like?} I want to become an anesthesiologist one day.} \\

\bottomrule
\end{tabular}%
}
\caption{\textbf{Example summaries for iterative preference optimization.} As the iterations progress, an increase in the extractiveness of the summaries is observed, with summaries increasingly incorporating sentences directly from the source. Sentences in \textbf{bold} indicate exact matches to the source text.}
\label{table:extractive_results}
\end{table*}

\subsection{License Information of The Assets Used in This Work}
\label{subsec:license}

\paragraph{Datasets} We report known license information of the assets used in this work. The following datasets used in this paper are under the MIT License: XSUM~\cite{xsum}. 
The following datasets used in this paper are under the CC BY 4.0 License: TL;DR~\cite{tldr}.

\paragraph{Models}
We report known license information of the assets used in this work. The following datasets used in this paper are under the Apache 2.0 License: GPT-J~\cite{gpt-j}, Mistral-7B~\cite{mistral}, BART~\cite{lewis2019bart}. The following datasets used in this paper are under the Llama2 License: LLaMA2-7B~\cite{touvron2023llama}

\paragraph{Source code}
We use the implementation of existing baseline methods for reporting their results in this paper. The source code utilized in this paper is subject to the MIT License: MINT~\cite{dreyer2023evaluating}, ITI~\cite{li2023inferencetime}, AlignScore~\cite{zha2023alignscore}, DoLa~\cite{chuang2023dola}, DCPMI~\cite{chae-etal-2024-mitigating}
The following source code utilized in this paper is subject to the BSD 3-Clause License: FactCC~\cite{kryscinski-etal-2020-evaluating}
The following source code utilized in this paper is subject to the CC-BY-NC-4.0 License: Lookahead~\cite{wan-etal-2023-faithful-generation}
The following source code utilized in this paper is subject to the Apache 2.0 License: BARTScore~\cite{yuan2021bartscore}, trl/examples/research\_projects/stack\_llama\_2~\cite{vonwerra2022trl}

\subsection{Statistics for Data}
\label{subsec:data_statistics}
We utilized two abstractive summarization datasets, TL;DR and XSUM. The TL;DR dataset is constructed by Reddit posts and their corresponding summaries, with 117k samples in the train split, 6.45k in the validation split, and 6.55k in the test split. The XSUM dataset consists of BBC articles and their corresponding summaries, totaling 204k samples in the train split, 11.3k in the validation split, and 11.3k in the test split. Both datasets are in English.

The train splits from each dataset were used during the SFT phase, the validation splits during the preference optimization phase, and the test splits during the evaluation phase.

\subsection{Analysis on Error Bars}
\label{subsec:error_bars}
All experiments were evaluated in single run, fixing the seed at 42. Additionally, all summary generations were conducted in the order of the provided test dataset.

\subsection{Reproducibility}
\label{supp:reproducibility}

We conducted our experiments using computing clusters equipped with NVIDIA RTX 6000 (GPU memory: 48GB) and NVIDIA RTX 3090 GPUs (GPU memory: 24 GB), allocating a single GPU for each experiment. 

Based on NVIDIA RTX 6000, model preference optimization typically required an average of 1 hour and 30 minutes. When generating summaries, using GPT-J (6B) with beam search decoding took approximately 20 hours, and with greedy decoding, about 5 hours and 30 minutes. Using Mistral-7B and LLaMA-7B models with beam search decoding took around 5 hours, while with greedy decoding, it took about 1 hour and 30 minutes.

\subsection{Parameters for Package}
\label{supp:package}
For evaluating summaries, we loaded ROUGE and BERTScore from the evaluate package (version: 0.4.1).


\end{document}